\documentclass[11pt,a4paper]{article}
\usepackage[hyperref]{acl2020}
\usepackage{times}
\usepackage{latexsym}

\usepackage{graphicx}
\usepackage{wrapfig}
\usepackage{makecell}
\usepackage{amsmath}
\usepackage{amssymb}
\usepackage{algorithm}
\usepackage{algpseudocode}
\usepackage{multirow}
\usepackage{url}
\usepackage{comment}

\usepackage{color}

\newenvironment{tight_itemize}{
\begin{itemize}
  \setlength{\itemsep}{0pt}
  \setlength{\parskip}{0pt}
}{\end{itemize}}
\usepackage{microtype}

\aclfinalcopy 

\title{Rethinking Exposure Bias in Adversarial Language Modeling}

\author{Yifan Xu \thanks{ \,\, Equal contribution.} \,, Kening Zhang $^{*}$, Haoyu Dong, Yuezhou Sun, Wenlong Zhao \& Zhuowen Tu   \\
University of California San Diego\\
\texttt{\{yix081,kez040, had002, yus174, wez094,ztu\}@ucsd.edu} \\
}

\date{}

\begin{document}
\maketitle
\begin{abstract}
Exposure bias refers to the phenomenon that a language model trained under the teacher forcing schema may perform poorly at the inference stage when its predictions are conditioned on its outputs that diverge from the training corpus. Although several adversarial training methods have been proposed to avoid teacher forcing, lacking a clear evaluation for the exposure bias remains a concern. The contribution of our work is two-fold. (1) We propose to evaluate exposure bias based on the quality of sentence generated in the sentence completion task. (2) We adopt two strategies, \textit{multi-range reinforcing} and \textit{multi-entropy sampling}, to stabilize adversarial training, and show an improvement over the competing models with regards to the sentence completion task and corpus BLEUs.
\end{abstract}

\vspace{-2mm}
\section{Introduction}
\vspace{-1mm}
Likelihood-based language models with deep neural networks have been widely adopted to tackle the language modeling tasks \citep{graves2013speech, karpathy2015deep, bahdanau2014neural}. By far, one of the most popular training strategies is \textit{teacher forcing}, which is derived from the general maximum likelihood estimation (MLE) principle \citep{williams1989learning}. Under the teacher forcing schema, the language model makes predictions conditioned on the ground-truth inputs. This is susceptible to so-called \textit{exposure bias}: a model may perform poorly at the inference stage, once its prefix diverges from the previously learned data \citep{bengio2015scheduled}. However, there is little work on how to expose and quantify such performance degeneration in text generation. 

A common strategy to mitigate the exposure bias problem is to impose additional supervision upon the model's self-generated output via adversarial training. 
The actor-critic (AC) method \citep{konda2000actor} and SeqGAN \citep{yu2017seqgan} introduce an additional critic network to offer rewards on a language model's self-generated sequences. Therefore, the language model can later, at the inference stage, predict robustly with its previous outputs. One issue in adversarial training is that the signal from the critic network is very sparse, which leads to stability issues. The second issue is about the non-stationary sampled data with strongly correlated online updates \citep{pfau2016connecting, mnih2016asynchronous}. Due to these problems, existing language GANs \citep{yu2017seqgan, lin2017adversarial, guo2017long} have a risk of compromising generation diversity \citep{caccia2018language}.
This paper makes the following contributions:
\vspace{0mm}

{\small
\vspace{-1mm}
\begin{enumerate}
\item We propose to evaluate the exposure bias for a language model by performing the sentence completion task using the ground truth prefix.
\vspace{-1mm}
\item We introduce a new approach, \textit{multi-entropy sampling} and \textit{multi-range reinforcing} (MEMR), to overcome the difficulties during adversarial training, which demonstrates a significant improvement over the competing models in the corpus BLEUs metrics, as well as our proposed measures in sentence completion.
\vspace{-1mm}
\end{enumerate}
}

\vspace{-2mm}
\section{Related Works}
\vspace{-1mm}
A common measure quantifying the exposure bias is still absent. Existing works often show performance gains by introducing adversarial training but questions remain if such gains indeed result in the reduction of the exposure bias \citep{bahdanau2016actor, yu2017seqgan}. Later works add generation diversity into consideration \citep{shi2018towards, caccia2018language, alihosseini2019jointly} or take a perspective from traditional language modeling aspects \citep{tevet2018evaluating}. A closely related work to our evaluation measure is \citep{he2019quantifying}. The difference is that \citet{he2019quantifying} requires inference for ground truth data distribution with experiments performed using synthetic data.

An early work addressing the exposure bias problem is \citep{bengio2015scheduled} in which a curriculum learning approach called \textit{scheduled sampling} is proposed by gradually replacing the ground-truth tokens with the model's predictions. In recent RL-inspired works, \citet{ranzato2015sequence} adopt the REINFORCE algorithm \citep{sutton2000policy} to directly optimize the test-time evaluation score. \citet{bahdanau2016actor} employ a similar approach by training a critic network to predict the metric score for the actor's generated sequence of tokens. In parallel, a language version of generative adversarial networks (GANs) \citep{goodfellow2014generative}, SeqGAN, is introduced in \citep{yu2017seqgan}. SeqGAN consists of a generator pre-trained under MLE and a discriminator pre-trained to discern the generator's distribution from the real data. Follow-up works such as RankGAN \citep{lin2017adversarial} and LeakGAN \citep{guo2017long} alter the training objectives or model architectures to enhance the guidance. RankGAN \citep{lin2017adversarial} replaces the binary reward with a relative ranking score. LeakGAN \citep{guo2017long} allows the discriminator to ``leak" its internal states to the generator at intermediate steps. \citet{shi2018towards} model a reward function using inverse reinforcement learning (IRL). 

\vspace{-1mm}
\section{Exposure Bias Evaluation}
\vspace{-2mm}
\subsection{Exposure Bias}
Cross-entropy loss adopted in teacher forcing is equivalent to minimizing the \textit{forward} KL divergence $D_{KL}(P||Q_\theta)$ between data distribution $P$ and model distribution $Q_\theta$. However, during the inference stage, the model is often evaluated based on the quality of its generated samples. The evaluation metrics or human experts can be seen as surrogates of the data distribution $P$, so what they measure is the \textit{reverse} KL divergence $ D_{KL}(Q_\theta||P)$.

In Bayesian inference, there is a well-known difference between $D_{KL}(P||Q)$ and $D_{KL}(Q||P)$ \citep{mackay2003information}. Minimizing $D_{KL}(P||Q)$ encourages the model to cover all the modes in the training data, which will result in over-generalization in the extreme case. In contrast, minimizing $D_{KL}(Q||P)$ prefers the model to concentrate on the largest mode while ignoring the others, which tends to cause mode collapse \citep{huszar2015not}.
In our language modeling task, an LSTM strives to cover the entire data distribution at the cost of over-generalization. It is more likely to produce prefixes different from those seen at the training stage, and the fact that this model has never learned to predict based on these prefixes potentially leads to the exposure bias . 

\subsection{Sentence Completion Task}
\vspace{-2mm}
In this section, we form a sentence completion task to evaluate the exposure bias. Given a sentence prefix $X_{1:k}$ of length K drawn from a data distribution $P$, we apply a language model $Q_\theta$ to perform sentence completion until final the $T$ step, starting from such prefix.

{\small 
\begin{tight_itemize}
\item  If the prefix $X_{1:k}$ is sampled from a seen distribution $P_{seen}$, then the exposure bias for the sentence completion task should be relatively low, where
\begin{equation}
{\scriptstyle  Q_\theta(X_{k:T}|P_{seen}) =  \mathbb{E}_{X_{1:k} \sim P_{seen}}  Q_{\theta}(X_{k:T}|X_{1:k}) }\nonumber
\vspace{-2mm}
\end{equation}
\item  If the prefix $X_{1:k}$ comes from an unseen data distribution $P_{unseen}$, then the exposure bias for the task can be critical, where
\begin{equation}
{\scriptstyle  Q_\theta(X_{k:T}|P_{unseen}) =  \mathbb{E}_{X_{1:k} \sim P_{unseen}}  Q_{\theta}(X_{k:T}|X_{1:k}) }\nonumber
\end{equation}
\end{tight_itemize}
\vspace{-2mm}
}

Based on the definition for the exposure bias, $Q_\theta(X_{k:T}|P_{unseen})$ should suffer more from the training-testing deviation than $Q_\theta(X_{k:T}|P_{seen})$. Also, such performance degeneration should be more significant when prefix $k$ grows longer in both scenarios. These two hypotheses are confirmed by our result in Figure \ref{fig:EMNLP_re_train}. 

As a measurement to assess model's generation quality, forward Corpus BLEU, $\textit{BLEU}_{\text{F}}$, is evaluated. Because precision is the primary concern, we set softmax temperature  $\tau=0.5$ to sample high-confidence sentences from model's distribution.  

Based on the task completion task results in Figure \ref{fig:EMNLP_re_train}, we observe that original SeqGAN \citep{yu2017seqgan} shows more stable result although many text GAN variants are proposed later, which is unexpected. Therefore, our method MEMR is motivated to improve SeqGAN by introducing denser reward signal from the critic network and further stabilizing the adversarial training.

\vspace{-1mm}
\section{Method Description}
\vspace{-1mm}
\subsection{Actor-Critic Training}
Actor-Critic methods (ACs) formulates language modeling as a generalized Markov Decision Process (MDP) problem, where the actor learns to optimize its policy guided by the critic, while the critic learns to optimize its value function based on the actor's output and external reward information. As \citet{pfau2016connecting} points out, GAN methods can be seen as a special case of AC where the critic aims to distinguish the actor's generation from real data and the actor is optimized in an opposite direction to the critic. 

In this work, we use a standard single-layer LSTM as the actor network. The training objective is to maximize the model's expected end rewards with policy gradient \citep{sutton2000policy}:
\vspace{-3mm}
\begin{equation}
{\scriptstyle  \mathcal{L}(\theta) = - \mathbb{E}_{X_{1:T}\sim \pi_{\theta}} \sum^{T}_{t=1}{Q_{\phi}(x_{t}, h_{t}) \log \pi_{\theta}(x_{t}|h_{t})}} \nonumber\\ 
\label{eq:ac_a_loss}
\vspace{-2mm}
\end{equation} 
\vspace{-1mm}
In practice, we perform a Monte-Carlo (MC) search with roll-out policy following \citet{yu2017seqgan} to sample complete sentences starting from each location in a predicted sequence and compute their end rewards. Empirically, we found out that the maximum, instead of average, of rewards in the MC search better represents each token's actor value and yields better results during training. Therefore, we compute the action value by:
\vspace{-4mm}
\begin{equation}
\vspace{-2mm}
{\scriptstyle Q_{\phi}(x_{t},h_{t}) = \max_{X_{t:T}\in MC^{\theta}(X_{1:t},T)}Q_\phi(X_{1:T})} \nonumber
\label{eq:MC search}
\vspace{-1mm}
\end{equation} 

Then, We use a convolutional neural network (CNN) as the critic to predict the expected rewards for current generated prefix: 
\vspace{-4mm}
\begin{equation}
\vspace{-2mm}
{\scriptstyle \mathcal{L}(\phi) = - \mathbb{E}_{X_{1:T}\sim \pi_{\theta}} (r(X_{1:T}) - {Q}_{\phi}(X_{1:T}))^2} \nonumber
\label{eq:lqr}
\end{equation} 

\subsection{MEMR}
During the experiment, we observe a certain level of instability for the learned models. In the previous literature, two major factors behind the training instability are the sparse reward from critic network and the update correlation in the sampling process \citep{pfau2016connecting, mnih2016asynchronous, volodymyr2013playing}. We address these problems using the following strategies:

\textbf{Multi-Entropy Sampling:} Language GANs can be seen as online RL methods, where the language model is updated from data generated by a single policy. Most sampled sentences in MC search are highly correlated. Similar to \citet{xu2019neural}, we empirically observe that increasing the range of the entropy of the actor's sample distribution during training is beneficial to the adversarial training performance. Specifically, we alternate the temperature $\tau$ in the softmax to generate samples under different behavior policies. During the critic's training, the ground-truth sequences are assigned a perfect target value of 1. The samples obtained with $\tau < 1$ are supposed to contain lower entropy, thus they receive a higher target value close to 1. Those samples obtained with $\tau > 1$ contain higher entropy, and the target value is closer to 0. This mechanism decorrelates updates during sequential sampling by sampling from multiple diverse entropy distributions synchronously. 

\textbf{Multi-Range Reinforcing:} Our idea of multi-range supervision takes inspiration from deeply-supervised nets (DSNs) \citep{lee2015deeply}. By design, lower layers in a CNN have smaller receptive fields, allowing them to make better use of local patterns. Differently from DSNs \citep{lee2015deeply} which disregard all intermediate predictions in the end, we average the reward predictions from multiple intermediate layers of the critic network with the final output, which attend to local n-grams rather than the whole complete sentence. This is a solution to the reward sparseness, as the language model can receive averaged reward with more local information.

\subsection{Effectiveness of Multi-Range Reinforcing and Multi-Entropy Sampling}
Table \ref{tab:ablationstudy} demonstrates the effectiveness of multi-entropy  sampling (ME) and multi-range reinforcing (MR). We observe that ME improves $\text{BLEU}_{\text{F5}}$ (precision) significantly while MEMR further enhances $\text{BLEU}_{\text{F5}}$ (precision) and $\text{BLEU}_{\text{F5}}$ (recall). Detailed explanations of these metrics can be found in Section \ref{evaluation}.\\

\begin{table}[!htb]
\vspace{-3mm}
\begin{center}
\vspace{-2mm}
\begin{small}
\scalebox{0.9}{
\begin{tabular}{lcc}
\hline
\hline
\textbf{Architecture} & $\textbf{BLEU}_{\textbf{F5}}$ & $\textbf{BLEU}_{\textbf{B5}}$ \\
\hline
\text{TF} & 15.4 $\pm$  $0.17$ & 30.5 $\pm$  0.08 \\
\hline
\text{AC} & 13.8 $\pm$ 0.16  & 30.3 $\pm$ 0.13  \\
\text{AC (with ME)} & 22.4 $\pm$ 0.25  & 30.0 $\pm$ 0.09 \\
\text{AC (with MEMR )} & 24.5 $\pm$ 0.14  & 31.6 $\pm$  0.10\\
\hline
\hline
\end{tabular}
}
\end{small}
\caption{\small{Effectiveness of the proposed ME and MEMR strategies on EMNLP2017 WMT News Dataset}}
\label{tab:ablationstudy}
\end{center}
\vspace{-7mm}
\end{table}

\begin{figure*}[!ht]
\begin{center}
 \vspace{-6mm}
\begin{tabular} {c}
 \includegraphics[width=0.4\textwidth ]{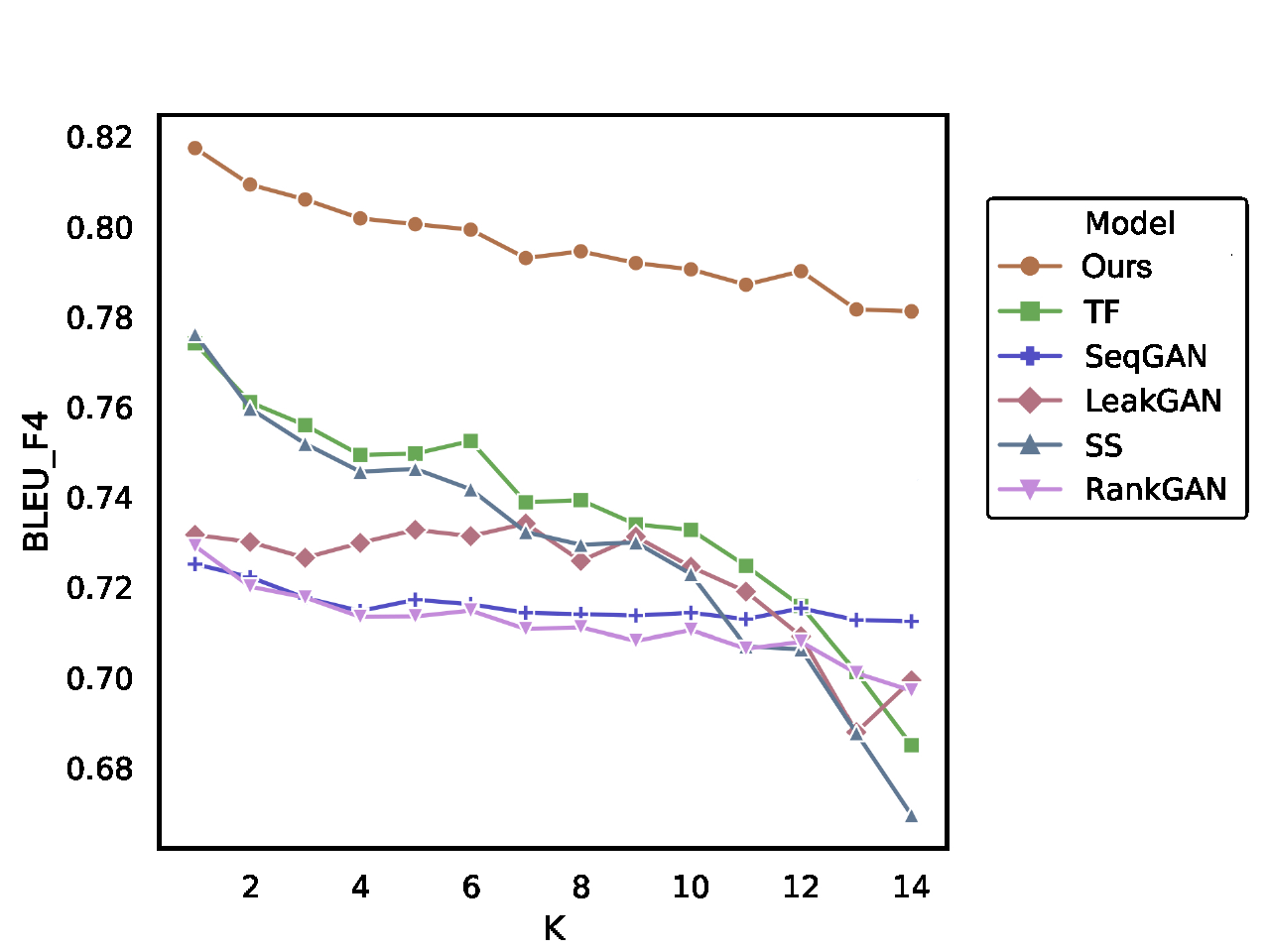} 
 \includegraphics[width=0.4\textwidth ]{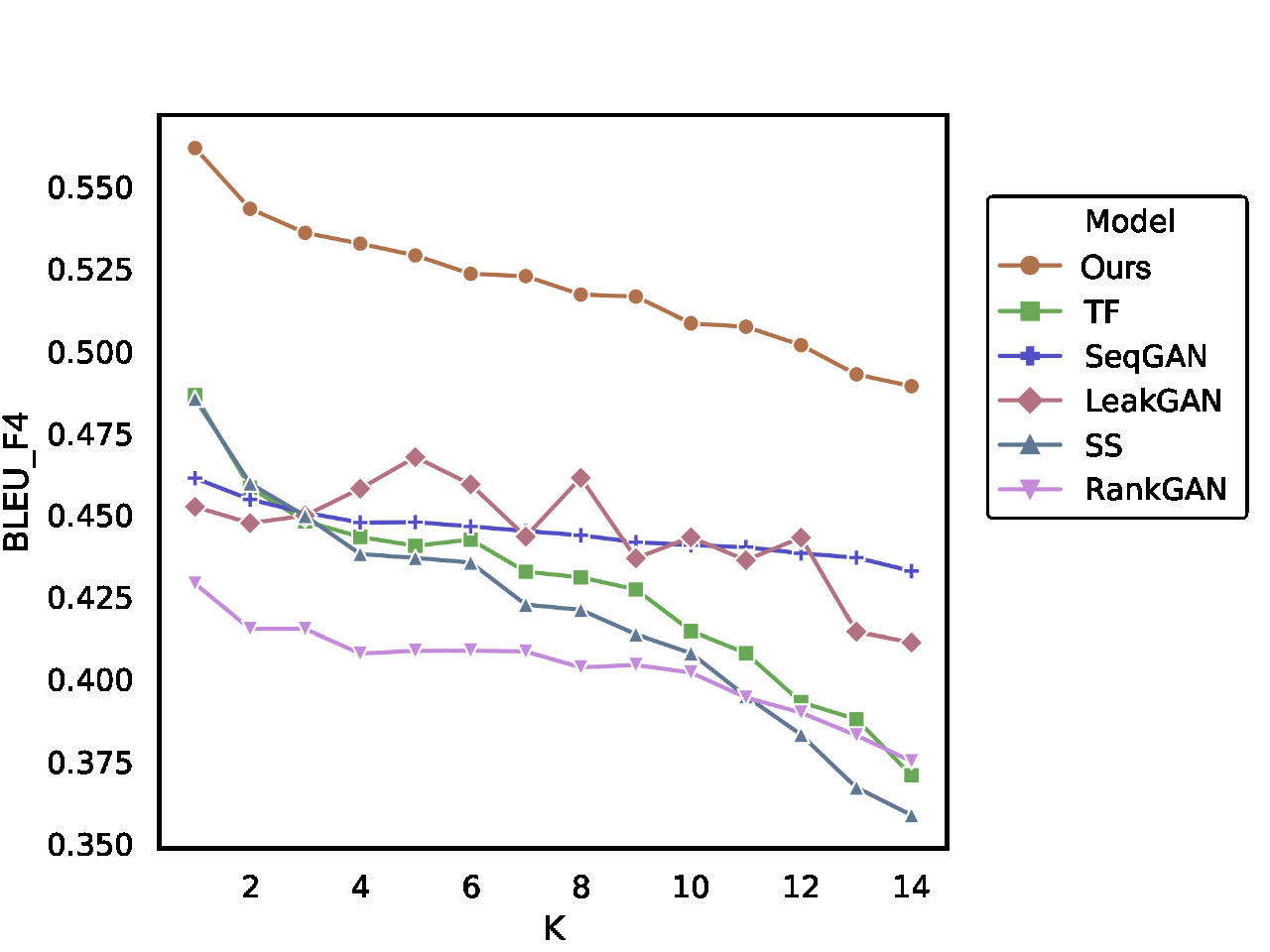} \\
 (a) Train data (Seen prefixes) \hspace{+5mm} (b) Test data (Unseen prefixes)
\end{tabular}
\vspace{-4mm}
\caption{\small{\textbf{Sentence Completion Task} results based on prefixes from training and testing datasets on EMNLP2017 WMT News [Higher is better]. In each experiment, the data source for the prefixes is used as the reference to calculate $\text{BLEU}_{\text{F4}}$.}}
\label{fig:EMNLP_re_train}
\end{center}
\end{figure*}

\begin{table*}[!htb]
\begin{center}
\vspace{-2mm}
\scalebox{0.7}{
\begin{tabular}{lccc|ccc}
\hline
\hline
  &  \multicolumn{3}{c|}{\textbf{EMNLP2017 WMT}} &  \multicolumn{3}{c}{\textbf{Google-small}} \\
\textbf{Model} & $\textbf{BLEU}_{\textbf{F5}}$  & $\textbf{BLEU}_{\textbf{B5}}$ & $\textbf{BLEU}_{\textbf{HA5}}$  & $\textbf{BLEU}_{\textbf{F5}}$  & $\textbf{BLEU}_{\textbf{B5}}$ & $\textbf{BLEU}_{\textbf{HA5}}$\\
\hline
\textsc{Teacher Forcing (TF)} & 15.4 $\pm$  0.11  & 30.5 $\pm$  0.05 & 20.5 $\pm$ 0.10 & 9.6 $\pm$  0.03 & 12.9 $\pm$  0.02 & 11.00 $\pm$  0.02 \\
\textsc{Scheduled Sampling (SS)} \citep{bengio2015scheduled} & 12.1 $\pm$ 0.14 & 30.3 $\pm$  0.06 & 17.3  $\pm$  0.14 & 6.2  $\pm$  0.04 &  10.7  $\pm$ 0.02 & 7.8  $\pm$ 0.04\\
\textsc{SeqGAN} \citep{yu2017seqgan} & 16.6 $\pm$  0.09& 28.7 $\pm$ 0.37 & 21.0 $\pm$ 0.11  &  20.7 $\pm$  0.02 & 14.4 $\pm$ 0.02 & 17.0 $\pm$  0.01\\
\textsc{RankGAN} \citep{lin2017adversarial} & 17.7 $\pm$  0.14 & 30.1 $\pm$ 0.06 & 22.3 $\pm$  0.11  & 21.4 $\pm$  0.06 & 12.7 $\pm$ 0.02 & 15.9 $\pm$ 0.02\\
\textsc{LeakGAN} \citep{guo2017long} & 19.8 $\pm$  0.11 & \textbf{31.6} $\pm$ 0.04 & 24.4 $\pm$  0.10 & -  &  - & - \\
\hline
\textsc{MEMR} (ours) &  \textbf{24.5} $\pm$ 0.08  &  \textbf{31.6} $\pm$ 0.06  &  \textbf{27.9} $\pm$ 0.07  &  \textbf{22.0} $\pm$  0.07 &  \textbf{15.8} $\pm$ 0.02  &  \textbf{18.4} $\pm$ 0.03 \\
\hline
\hline
\end{tabular}
}
\caption{\small Corpus BLEUs Results on EMNLP2017 WMT News and the Google-small dataset. The 95 \% confidence intervals from multiple trials are reported. {\scriptsize \textsuperscript{$\dagger$} the Google-small was not tested in \citep{guo2017long} and we are unable to train LeakGAN on this dataset using the official code due to its training complexity (taking 10+ hours per epoch).}}
\label{tab:EMNLP}
\end{center}
\vspace{-8mm}
\end{table*}

\section{Experiment}
\label{experiment}
\vspace{-2mm}
\subsection{Datasets}
We perform evaluations on two datasets: \textit{EMNLP2017 WMT News} \footnote{https://github.com/geek-ai/Texygen} and \textit{Google-small}, a subset of Google One Billion Words \footnote{http://www.statmt.org/lm-benchmark/}. 
\vspace{-2mm}
{\small 
\begin{itemize}
\item \textit{EMNLP2017 WMT News} is provided in \citep{zhu2018texygen}, a benchmarking platform for text GANs. The entire dataset is split into a training set of 195,010 sentences, a validation set of 83,576 sentences, and a test set of 10,000 sentences. The vocabulary size is 5,254 and the average sentence length is 27. 
\item \textit{Google-small} is sampled and pre-processed from the Google One Billion Words. It contains a training set of 699,967 sentences, a validation set of 200,000 sentences, and a test set of 99,985 sentences. The vocabulary size is 61,458 and the average sentence length is 29. 
\end{itemize}
}
\vspace{-3mm}

\subsection{BLEU metric}
\label{evaluation}
We adopt three variations of BLEU metric from \citet{shi2018towards}. 
$\textit{BLEU}_{\textit{F}}$, or forward BLEU, is a metric for precision, and $\textit{BLEU}_{\textit{B}}$, or backward BLEU, is a metric for recall. $\textit{BLEU}_{\textit{HA}}$ computes the harmonic mean of both BLEU. These three metrics take both diversity and quality into consideration. A model with severe mode collapse or diverse but incorrect outputs receives low scores.\\
\vspace{-4mm}

\subsection{Implementation Details}
\vspace{-1mm}
We implement a standard single-layer LSTM as the generator (actor) and a eight-layer CNN as the discriminator (critic). The LSTM has embedding dimension 32 and hidden dimension 256. The CNN consists of 8 layers with filter size 3, where the 3rd, 5th, and 8th layers are directly connected to the output layer for  multi-range supervision. Other parameters are consistent with \citet{zhu2018texygen}. Adam optimizer is deployed for both critic and actor with learning rate $10^{-4}$ and $5 \cdot 10^{-3}$ respectively. The target values for the critic network are set to [0, 0.2, 0.4, 0.6, 0.8] for samples generated by the LSTM with softmax temperatures [0.5, 0.75, 1.0, 1.25, 1.5].

 \vspace{-2mm}
\section{Results}
 \vspace{-2mm}
\label{sec:eb}
Based on the sentence completion results in Figure \ref{fig:EMNLP_re_train}, all models decrease in precision of generated text (reflected via $\text{BLEU}_{\text{F4}}$) as the fed-in prefix length ($K$) increases, but the effect is stronger on the unseen test data, revealing the existence of exposure bias. Nonetheless, our model trained under ME and MR yields the best sentence quality and a relatively moderate performance decline. 

Although TF and SS demonstrate higher $\text{BLEU}_{\text{F5}}$ performance with shorter prefixes, their sentence qualities drop drastically on the test dataset with longer prefixes. On the other hand, GANs begin with lower $\text{BLEU}_{\text{F4}}$ precision scores but demonstrate less performance decay as the prefix grows longer and gradually outperform TF. This robustness against unseen prefixes exhibits that supervision from a learned critic can boost a model's stability in completing unseen sequences. The better generative quality in TF and the stronger robustness against exposure bias in GANs are two different objectives in language modeling, but they can be pursued at the same time. Our model's improvement in both perspectives exhibit one possibility to achieve the goal.

We also report Corpus BLEUs to reflect the quality and diversity of generated text in Table \ref{tab:EMNLP} with competing models on EMNLP2017 WMT News and Google-small. Our model, MEMR, outperforms the others in Corpus BLEUs, indicating a high diversity and quality in its sample distribution. 

 \vspace{-2mm}
\section{Conclusion}
\vspace{-2mm}
We propose to use the sentence completion task to reveal exposure bias in text generation. Further, we overcome the hurdles in adversarial training with \textit{multi-range reinforcing}  and \textit{multi-entropy sampling} (MEMR), which shows an improvement in the sentence completion task and Corpus BLEUs.

\subsubsection*{Acknowledgments}
The authors are grateful for the supports by NSF IIS-1618477, NSF IIS-1717431, and a grant from Samsung Research America.
\bibliography{anthology,acl2020}
\bibliographystyle{acl_natbib}

\appendix

\end{document}